%% file: samplepaper.tex
\documentclass[runningheads]{llncs}
\usepackage[T1]{fontenc}
\usepackage{graphicx}
\usepackage{subfig}
\usepackage{comment}
\usepackage{amsmath}
\usepackage{amssymb}
\usepackage{xcolor}
\usepackage{tikz}
\usetikzlibrary{arrows}
\usetikzlibrary{arrows.meta}

\DeclareMathOperator*{\argmax}{arg\, max}
\newcommand{\ada}{\text{ADA}^*}
\newcommand{\auc}{\text{AuROC}}
\newcommand{\tpr}{\text{TPR}_{5\%}}

\definecolor{aliceblue}{rgb}{0.94, 0.97, 1.0}
\definecolor{honeydew}{rgb}{0.94, 1.0, 0.94}
\definecolor{bananamania}{rgb}{0.98, 0.91, 0.71}
\definecolor{lavenderblush}{rgb}{1.0, 0.94, 0.96}

\usepackage[pagebackref=true,breaklinks=true,colorlinks,bookmarks=false]{hyperref}

\usepackage{color}

\begin{document}
\title{Detecting Adversarial Attacks in Semantic Segmentation via Uncertainty Estimation: A~Deep~Analysis}
\titlerunning{Detecting Adversarial Attacks in Semantic Segmentation}
%
\author{Kira Maag\inst{1}\orcidID{0000-0003-1767-0476} \and Roman Resner\inst{2} \and
Asja Fischer\inst{2}\orcidID{0000-0002-1916-7033} }
\authorrunning{K. Maag et al.}
%
\institute{Technical University of Berlin, Berlin, Germany \and
Ruhr University Bochum, Bochum, Germany\\
\email{maag@tu-berlin.de, \{roman.resner, asja.fischer\}@rub.de}}
\maketitle              
\begin{abstract} 
Deep neural networks have demonstrated remarkable effectiveness across a wide range of tasks such as semantic segmentation. Nevertheless, these networks are vulnerable to adversarial attacks that add imperceptible perturbations to the input image, leading to false predictions. This vulnerability is particularly dangerous in safety-critical applications like automated driving. While adversarial examples and defense strategies are well-researched in the context of image classification, there is comparatively less research focused on semantic segmentation. Recently, we have proposed an uncertainty-based method for detecting adversarial attacks on neural networks for semantic segmentation \cite{visapp24}. We observed that uncertainty, as measured by the entropy of the output distribution, behaves differently on clean versus adversely perturbed images, and we utilize this property to differentiate between the two. In this extended version of our work, we conduct a detailed analysis of  uncertainty-based detection of adversarial attacks including a diverse set of adversarial attacks and various state-of-the-art neural networks. Our numerical experiments show the effectiveness of the proposed uncertainty-based detection method, which is lightweight and operates as a post-processing step, i.e., no model modifications or knowledge of the adversarial example generation process are required.

\end{abstract}
%
%
%
%
%
\section{Introduction}
\begin{figure}[t]
    \centering
    \subfloat{\includegraphics[width=0.329\textwidth]{
    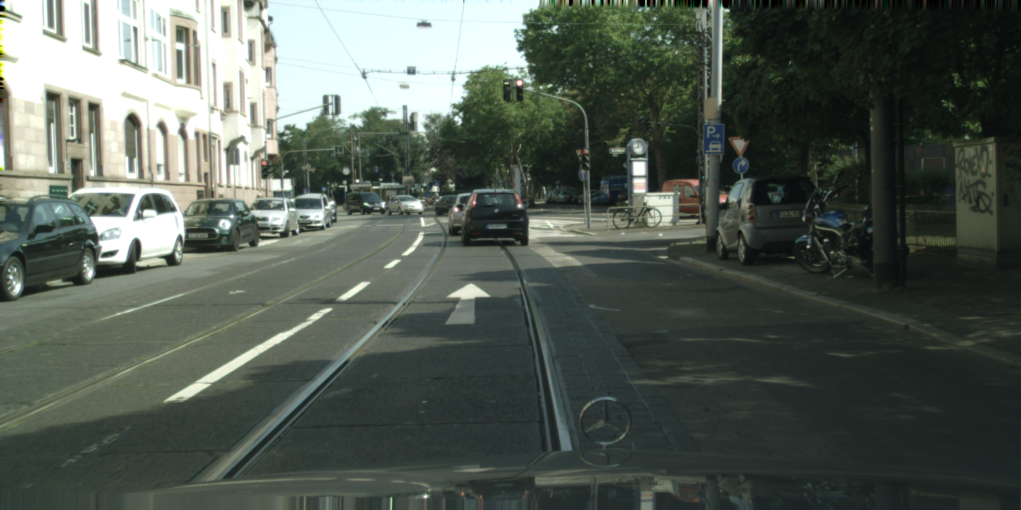} }
    \subfloat{\includegraphics[width=0.329\textwidth]{
    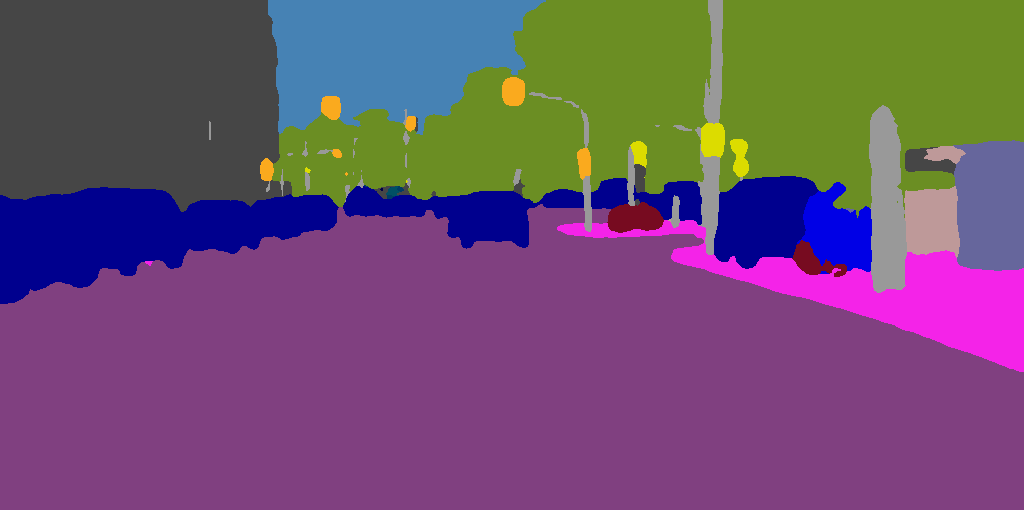} }
    \subfloat{\includegraphics[width=0.329\textwidth]{
    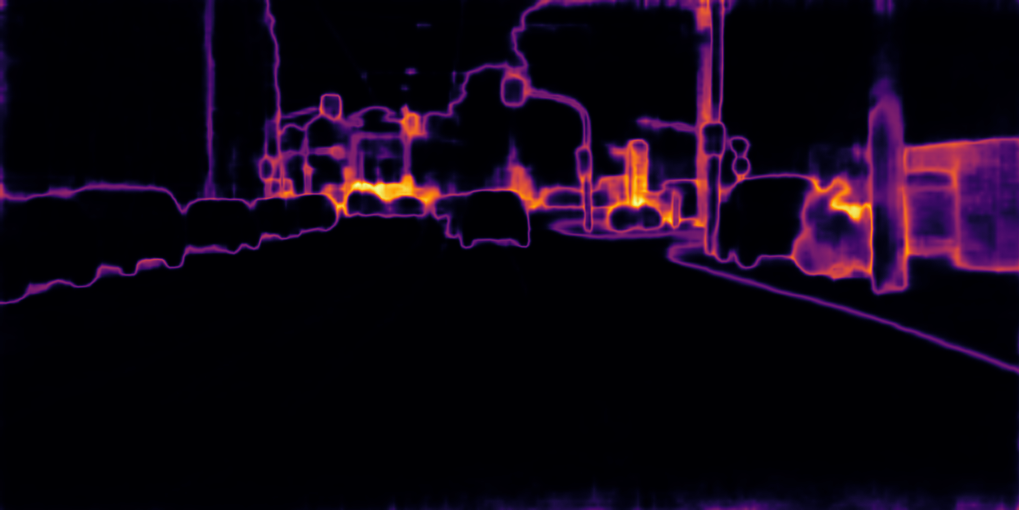} } \\
    \vspace{-2ex}
    \setcounter{subfigure}{0}
    \subfloat[][Input image]{\includegraphics[width=0.329\textwidth]{
    figs/frankfurt_000000_002963_rgb.png} }
    \subfloat[][Semantic segmentation]{\includegraphics[width=0.329\textwidth]{
    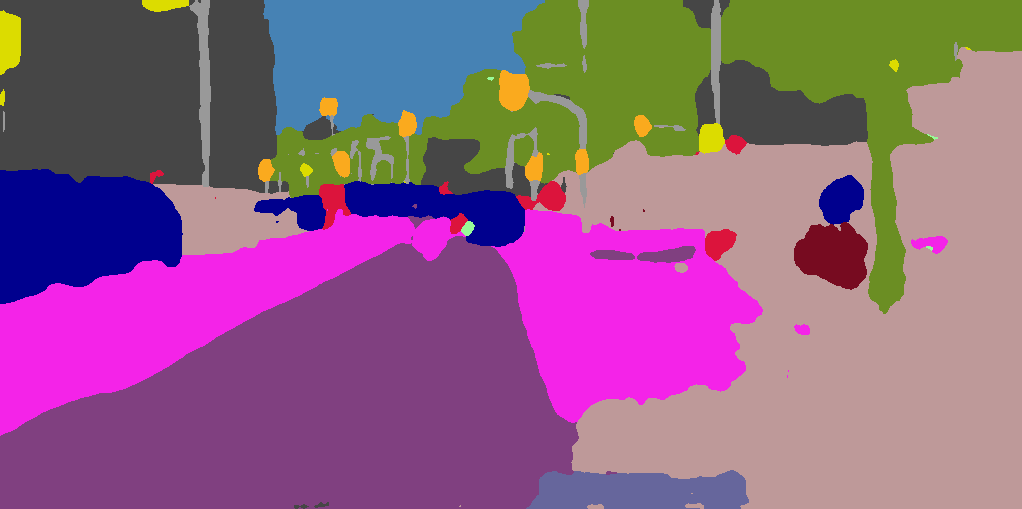} }
    \subfloat[][Entropy heatmap]{\includegraphics[width=0.329\textwidth]{
    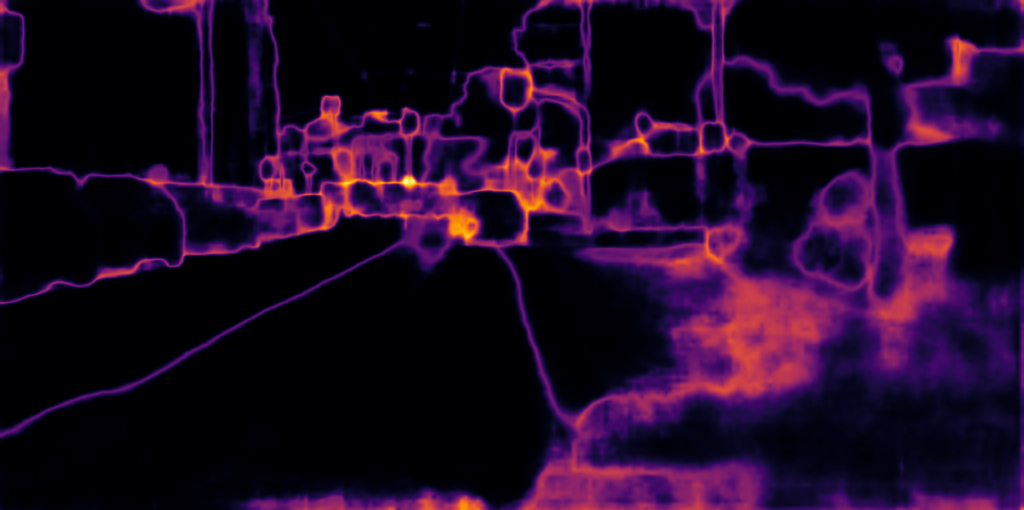} }
    \caption{Semantic segmentation prediction and entropy heatmap for a benign image (\emph{top row}) and adversarial example created by the FGSM attack \cite{Goodfellow2015} from that image (\emph{bottom row}).} 
    \label{fig:pred_heat}
\end{figure}

In recent years, deep neural networks (DNNs) have exhibited remarkable performance and demonstrated high expressiveness across a wide array of tasks, like semantic image segmentation \cite{Pan2022,Xu2023}. Semantic segmentation corresponds to segmenting objects in an image by assigning each pixel to a predefined set of semantic classes, thereby offering detailed and precise information about the scene. Despite their success, DNNs are vulnerable to \emph{adversarial attacks} \cite{Bar2021,Maag2024}, which pose significant risks in safety-critical applications such as automated driving. Adversarial attacks add small perturbations to input images, leading the DNN to make incorrect predictions. These perturbations are imperceptible to humans, making the detection of such attacks highly challenging, as illustrated in the first column of Fig.~\ref{fig:pred_heat} that shows a benign image and its adversary counterpart. 
The vulnerability caused by adversarial examples is a major security concern for real-world applications. Therefore, developing effective strategies to counter adversarial attacks is crucial. These approaches can either enhance the robustness of DNNs, making it harder to create adversarial examples (\emph{defense} approaches) or focus on detecting the presence of adversarial attacks (\emph{detection} approaches).

The research around adversarial attacks has garnered significant attention, which lead to the proposal of numerous attack and defense/detection strategies \cite{Khamaiseh2022}. However, most  adversarial example research has been confined to standard image classification models, typically using small datasets like MNIST \cite{LeCun2010} or CIFAR10 \cite{Krizhevsky2009}. 
The vulnerability of DNNs to adversarial attacks in more complex tasks such as semantic segmentation, particularly on real-world datasets from various domains, remains underexplored. The adversarial attacks on semantic segmentation networks can be broadly categorized into three categories. 
The first category consists of common attacks that were directly transferred from image classification to semantic segmentation models, making use of the fact that segmentation corresponds to pixel-wise classification \cite{Agnihotri2023,Gu2022,Rony2022}.
The second category includes attacks specifically designed for semantic segmentation, either causing the network to predict a predefined and image-unrelated segmentation mask \cite{Cisse2017,Xie2017} or completely omitting a segmentation class \cite{Metzen2017} (e.g.\ passengers in street scenes). These attacks are generally more challenging to detect compared to those that perturb pixels independently. 
The third category comprises attacks that generate smaller rectangular patches within the input, leading to incorrect predictions for the whole image \cite{Nakka2020,Nesti2022}.
Defense methods aim to achieve robustness against such attacks, such that high prediction accuracy is also obtained on perturbed images. For semantic image segmentation, defense approaches are often effective against only a single type of attack \cite{Arnab2018,Klingner2020,Yatsura2022}. 
Detection methods focus on classifying inputs as either benign or adversely perturbed, e.g.\ based on the segmentation model's output \cite{Xiao2018}.

In this work, we conduct a deep analysis of our uncertainty-based detection approach to distinguish between benign samples and adversarial attacks on semantic segmentation models. We investigate the detection performance of the proposed approach for various types of attacks and recent network architectures such as transformers. 
Uncertainty information has been utilized for detecting adversarial attacks on DNNs for classification before. 
Feinman et al.\ \cite{Feinman2017} proposed using Monte-Carlo Dropout, an approximation of Bayesian inference, to estimate model uncertainty for detecting adversarial attacks. By Michel et al.\ \cite{Michel2022} a gradient-based approach was developed to generate salient features for training a detector. Both methods require internal model access. 
While explored in the image classification context, uncertainty-based detection of adversarial attacks on segmentation models was previously unexplored. Moreover, in  contrast to the previously proposed uncertainty-based approaches, our approach operates as a post-processing step, relying solely on network output information. Specifically, we derive features from the uncertainty information provided by the DNN, such as the entropy of the pixel-wise output distributions \cite{Maag2020}. 
Figure~\ref{fig:pred_heat} (c) shows entropy heatmaps for a benign image (top) and a adversary image (bottom), highlighting high uncertainties in successfully attacked regions, thereby supporting the use of uncertainty information to distinguish between benign images and adversarial examples. 
On the one hand, we aggregate these pixel-wise uncertainty measurements over the images and use (i) the resulting aggregated quantities of benign images for training a one-class support vector machine for unsupervised novelty detection \cite{Weerasinghe2018} and (ii) the aggregated quantities of both benign and adversely perturbed images to train a  logistic regression model for classification. 
On the other hand, we augment this classification approaches on aggregated uncertainty measures by investigating classification based on the entire heatmaps (i.e., all pixel-wise uncertainty estimates), that is we investigate the question if the pixel-wise  information improves the classification. 
The adversarial examples used for training are generated by a single adversarial attack method (that is not used for evaluation), while the detector is designed to be applied to adversarial examples from various methods.
Our approach does not modify the semantic segmentation model or require knowledge of the adversarial example generation process. We only assume our post-processing model remains private, even if the attacker has full access to the semantic segmentation model.

We summarize our contributions in this work as follows:
\begin{itemize}
    \item We study our detection method that is not tailored to a specific type of adversarial attack but instead demonstrates high detection capability across various types. Here, we consider various attacks that have been developed for classification or semantic segmentation including untargeted and targeted attacks.
    \item We conduct a detailed analysis on the robustness of different semantic segmentation architectures, such as convolutional and transformer networks, with respect to adversarial attacks.
    \item Through a comprehensive empirical analysis, we demonstrate that uncertainty measures can effectively differentiate between clean and perturbed images. Our method achieves an average optimal detection accuracy rate of $89.36\%$ over the different attacks and network architectures.
\end{itemize}
%
%
%
\section{Related Work}\label{sec:rel_work}
In this section, we review related works on defense and detection methods for the semantic segmentation task. Detection methods classify model inputs as either benign or malicious, while defense methods aim to maintain high prediction accuracy even on perturbed images, thereby increasing the robustness of the model.
Adversarial training approaches enhance model robustness, such as the dynamic divide-and-conquer strategy \cite{Xu2021} and multi-task training \cite{Klingner2020}, which extends supervised semantic segmentation with self-supervised monocular depth estimation using unlabeled videos.
Another defense strategy involves input denoising to remove perturbations without retraining the model. Techniques like image quilting and the non-local means algorithm are presented by Bär et al.\ \cite{Bar2021} as data denoising methods.
A denoising autoencoder is used by Cho et al.\ \cite{Cho2020} to restore the original image and denoise the perturbation. 
The demasked smoothing technique from Yatsura et al.\ \cite{Yatsura2022} reconstructs masked regions of images using an inpainting model to defend against patch attacks. 
Improving model robustness during inference is another defense strategy. 
Mean-field inference and multi-scale processing, as investigated by Arnab et al.\ \cite{Arnab2018}, naturally form an adversarial defense.
The non-local context encoder proposed by He et al.\ \cite{He2019} models spatial dependencies and encodes global contexts to strengthen feature activations, fusing multi-scale information from pyramid features to refine predictions and generate segmentation. 
While these methods primarily focus on improving model robustness, there is limited work on detecting adversarial attacks on segmentation models. To the best of our knowledge, the only existing detection method is the patch-wise spatial consistency check introduced by Xiao et al.\ \cite{Xiao2018}.

The defense approaches described are designed for and tested against specific types of attacks. The issue with this is that while they show high model robustness with respect to the specific attacks, these defense methods might perform poorly against new, unseen attacks. 
In contrast, we study an uncertainty-based detection approach that shows strong results across various types of adversarial attacks.
Unlike the detection method presented by Xiao et al.\ \cite{Xiao2018}, which is tested only on perturbed images manipulated to predict a specific image and relies on randomly selecting overlapping patches to obtain pixel-wise confidence vectors, our approach uses information from a single network output inference. This avoids the computational expense of multiple network runs. 

In comparison to our first publication \cite{visapp24}, where the methodological focus is on adversarial example detection based on aggregated uncertainty measures over the image, in this paper we additionally investigate the merit value when a classifier is trained on the full heatmap. Moreover, we apply our detection method to further and more recent attacks as well as network architectures (e.g.\ transformers).
%
%
%
\section{Adversarial Attacks on Semantic Segmentation Models}\label{sec:attacks}
\paragraph{Semantic segmentation.}
Given an input image $x$, the semantic segmentation, i.e., classification of image content on pixel-level, is obtained by assigning a label $y$ from a prescribed label space $C = \{y_{1}, \ldots, y_{c} \}$ to each pixel $z$.
A neural network (with learned weights $w$) predicts for the $z$-th pixel a probability distribution specified by a probability vector 
$f(x;w)_{z} \in [0,1]^{|C|}$, that collects the probability $p(y|x)_z$ 
for each class $y \in C$.
The predicted class is then calculated by $\hat y_{z}^x =\argmax_{y\in\mathcal{C}} p(y|x)_z$.
The semantic segmentation network is trained on a pixel-wise loss function (typically the cross entropy loss) which is simultaneously minimized for all pixels $z \in Z$ of an image $x$.
The loss function per image is thus defined by
\begin{equation}\label{eq:loss}
    L(f(x;w),y) = \frac{1}{|Z|} \sum_{z \in Z} L_{z}(f(x;w)_{z},y_{z})  \enspace,
\end{equation} 
where $y_{z}$ is given as one-hot encoding.
%
%
\paragraph{Adversarial Attacks.}
A well-known attack on classification models, which is also frequently used for attacking semantic segmentation models, is the \emph{fast gradient sign method} (FGSM, \cite{Goodfellow2015}).
In the untargeted case (in which the attacker aims at predicting any wrong class), this single-step attack adds a small perturbation to the image $x$ leading to an increase of the loss defined in eq.~\eqref{eq:loss}. This is done by moving each pixel of the image a tiny step into the direction of the sign of the corresponding derivative of the loss function with respect to that pixel. Thus, the adversarial example is given by
\begin{equation} \label{eq:fgsm}
x^{\mathit{adv}} = x + \varepsilon \cdot \text{sign}(\nabla_x L(f(x;w),y)) \enspace, 
\end{equation}
where $\varepsilon$ describes the magnitude of perturbation, i.e., the $\ell_{\infty}$-norm of the perturbation is bounded to be (at most) $\varepsilon$. 
In the targeted case (in which the attacker aims at predicting a specific incorrect target label), this attack decreases the loss for the target label $y_{ll}$, that is, the adversarial example is given by 
\begin{equation}\label{eq:fgsm_ll}
    x^{\mathit{adv}} = x - \varepsilon \cdot \text{sign}(\nabla_x L(f(x;w),y^{ll})) \enspace . 
\end{equation}
Following the convention, the least likely class predicted by the segmentation model is chosen as target.
This attack is extended to the \emph{iterative FGSM} (I-FGSM, \cite{Kurakin2017}) increasing the attack strength by performing multiple steps of (clipped) gradient-based updates. 
In the untargeted case, it is given by \begin{equation} \label{eq:ifgsm}
    x^{\mathit{adv}}_{t+1} = \text{clip}_{x,\varepsilon} (x^{\mathit{adv}}_{t} + \alpha \cdot \text{sign}(\nabla_{x^{\mathit{adv}}_{t}} L(f(x^{\mathit{adv}}_{t};w),y))) \enspace,
\end{equation}
where $x^{\mathit{adv}}_{0} = x$, $\alpha$ is the step size, and $clip_{x,\epsilon}(\cdot)$ is a clip function ensuring that $x^{\mathit{adv}}_{t} \in [x-\varepsilon, x+\varepsilon]$. 
The targeted attack can be formulated analogously. 
The \emph{projected gradient descent} (PGD, \cite{Madry2018}) attack is similar to the iterative FGSM. The key difference is that PGD performs the update in the gradient direction and does not only consider the signs of the single derivatives. 
These methods serve as the foundation for more advanced attacks, such as \emph{orthogonal PGD} \cite{Bryniarski2022} and \emph{DeepFool} \cite{Moosavi2016}.

Additionally, specific adversarial attacks have been developed for the semantic segmentation task, including adaptations of the PGD attack \cite{Agnihotri2023,Gu2022}. 
A further advancement is the \emph{ALMA prox} attack \cite{Rony2022}, which produces adversarial perturbations with much smaller $\ell_{\infty}$-norm in comparison to FGSM and PGD using proximal splitting.
A \emph{certified radius-guided} (CR) attack framework for semantic segmentation models is introduced in \cite{Qu2023}. The certified radius defines the size of an $\ell_p$ ball around a pixel within any perturbation is guaranteed not to alter the predicted class for that pixel. The framework aims to target and disrupt pixels with relatively smaller certified radii as a larger certified radius indicates greater robustness to adversarial perturbations. 
Another weighting scheme for the loss function was introduced in \cite{Maag2024} where pixel classifications that are more easily perturbed are weighted higher and the pixel-wise losses corresponding to those pixels that are already confidently misclassified are zero-out.

Choosing the least likely class as the target, this results in network predictions appearing unrealistic and completely broken. 
For this reason, targeted attacks were developed (especially for the image segmentation task), whose prediction is similar to that of a clean image such as the \emph{Dense Adversary Generation} (DAG) attack \cite{Xie2017} or the \emph{stationary segmentation mask method} (SSMM) \cite{Cisse2017,Metzen2017}. The pixels of an image are iteratively perturbed until the majority of them are misclassified as belonging to the target class defined by the attacker's arbitrary segmentation.
Also aiming at a misclassification into a specific predefined segmentation, 
the universal perturbation is introduced in \cite{Metzen2017} achieving real-time performance for the attack at test time. The universal noise is learned using training data so that the desired target for unseen images is obtained during test time with same fixed noise applied to the images.

Another targeted attack is the \emph{dynamic nearest neighbor method} (DNNM) \cite{Chen2022,Metzen2017}. 
The goal is to remove one desired target class (like pedestrians or cars from street scene images) but keep for all other classes the network’s segmentation unchanged.
Instead of  adding noise to all pixels of an image as done by all attacks described so far, \emph{patch attacks} \cite{Nakka2020,Nesti2022} aim to disturb only a rectangular image area, but change the whole prediction.
Note, all of these attacks belong to the white box setting, i.e., for all of these attacks the adversarial attacker has full access to the model, including parameters and the loss function used for training.
%
%
%
\section{Uncertainty-based Detection Method}\label{sec:method}
The degree of uncertainty in a semantic segmentation prediction is quantified by dispersion measures on pixel-level.
An often used uncertainty measure is the \emph{entropy} which is related to a weighted mean probability margin and defined via 
\begin{equation}
    E(x)_{z} =-\frac{1}{\log(|C|)} \sum_{y\in \mathcal{C}}p(y|x)_{z} \cdot \log p(y|x)_{z} \enspace.
\end{equation}
The entropy heatmaps for a clean image (top) and a perturbed image (bottom) are shown in Fig.~\ref{fig:pred_heat} (c). These heatmaps reveal that higher uncertainties are concentrated in the attacked regions of the perturbed image. This observation underscores the potential of using uncertainty information as a tool for distinguishing between clean and perturbed data, providing a clear motivation for leveraging such information in our detection approach.
Moreover, the \emph{variation ratio}
\begin{equation}
    V(x)_{z} = 1 - p(\hat y_{z}^x|x)_z \enspace
\end{equation}
and the \emph{probability margin}
\begin{equation}
    M(x)_{z} = p(\hat y_{z}^x|x)_z - \max_{y\in\mathcal{C}\setminus\{\hat y_{z}^x\}} p(y|x)_{z} 
\end{equation}
provide information about the uncertainty of the prediction (in terms of the sum of probabilities for all non-predicted classes and distance to the second most likely class).

\begin{figure*}[t]
    \centering
    \scalebox{0.82}{\input{figs/method.tex}}
    \caption{Schematic illustration of our detection method, based on the figure from \cite{visapp24}. In the withe box setting, the attacker has full access to the semantic segmentation model. Uncertainty heatmaps are obtained based on the network output and these are used as input (either unfiltered or aggregated over the images) for the detection model classifying between clean and perturbed images.}
    \label{fig:method}
\end{figure*}
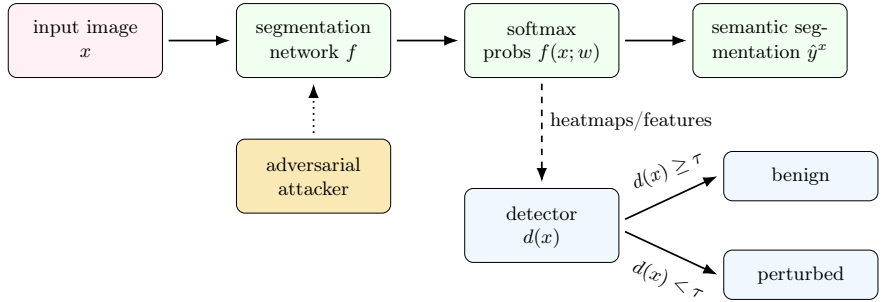

Formally, the detection task our approach targets complies with the classification between $d(x) < \tau$, i.e., image $x$ is adversely perturbed, and $d(x) \geq \tau$, i.e., $x$ is benign. Here, $\tau$ is a predefined detection threshold and $d(x)$ the probability of the given image $x$ to be benign  that is provided by a classification model. 
An overview of our approach is shown in Fig.~\ref{fig:method}. 

We study different ways to construct such a classifier.
On the one hand, we directly feed the entropy uncertainty heatmaps into a shallow neural network with two convolutional layers. 
On the other hand, we construct image-wise features from these pixel-wise dispersion measures as input for the classification models. To this end, we compute the averages $\bar D = 1 / |Z|\sum_{z \in Z} D(x)_{z}$, $D \in \{E,V,M\}$, as well as additionally consider the mean class probabilities for each class $y \in C$.
In total, we thus obtain $|C|+3$ features serving as input for the different classifiers.
Firstly, we use a basic uncertainty-based detector by thresholding only on the mean entropy $\bar{E}$ which merely requires selecting an appropriate threshold value.
Secondly, we examine two basic outlier detection techniques that only require benign data: a one-class support vector machine (OCSVM, \cite{Scholkopf1999}) and an approach for detecting outliers in a Gaussian distributed dataset by learning the ellipsoid region that results from thresholding the Gaussian at a chosen quantile fitted to the training data \cite{Rousseeuw1999}. 
Thirdly, we use a supervised logistic regression model (LASSO, \cite{Tibshirani1996}) trained on features extracted from both benign images and adversarial attacks. 
Note, the detection method is lightweight, i.e., the computation of the dispersion measure is inexpensive and the classifiers are trained before running the inference. 
%
%
%
\section{Numerical Experiments}\label{sec:exp}
We present the experimental setting first and then evaluate the adversarial attack performance as well as the detection capability of our detection method.
%
%
\subsection{Experimental Setting}\label{sec:exp_setting}
%
%
\paragraph{Dataset.} 
We conduct our studies on the Cityscapes \cite{Cordts2016} dataset representing the street scenario of dense urban traffic in various German cities. This dataset contains $2,\!975$ training and $500$ validation images of $18$ and $3$ different towns, respectively.
Cityscapes is often used to study adversarial attacks in semantic segmentation and represents more realistic scenarios than other datasets that contain only one or a few objects per image.
%
%
\paragraph{Segmentation Models.} 
We consider various state-of-the-art semantic segmentation networks. On the one hand, we use convolutional neural networks such as PIDNet \cite{Xu2023}, DDRNet \cite{Pan2022} and DeepLabv3+ \cite{Chen2018}, and on the other hand, transformer-based architectures like SETR \cite{Zheng2021} and SegFormer \cite{Xie2021}. The mean intersection over union (mIoU) values for these networks trained and evaluated on the Cityscapes dataset are given in Table~\ref{tab:miou}.
\begin{table}[t]
\caption{mIoU values on the Cityscapes validation set of the different semantic segmentation models.}
\centering
\scalebox{0.99}{
\begin{tabular}{ccccc}
\cline{1-5}
\phantom{m} PIDNet \phantom{m} & \phantom{m} DDRNet \phantom{m} & \phantom{m} DeepLabv3+ \phantom{m} &  \phantom{m} SETR \phantom{m} & \phantom{m} SegFormer \phantom{m} \\
\cline{1-5}
$80.89$ & $79.99$ & $80.21$ & $77.00$ & $82.25$ \\ 
\cline{1-5}
\end{tabular} }
\label{tab:miou}
\end{table} 
%
%
\paragraph{Adversarial Attacks.} 
For our experiments, we investigate attacks originally developed for classification models (FGSM, I-FGSM and PGD), as well as wide range of attacks for semantic segmentation models (ALMA prox, DAG, SSMM and DNNM).
For the FGSM and I-FGSM attacke, we use the parameter setting introduced in \cite{Kurakin2017}: the step size is given by $\alpha=1$, the perturbation magnitude by $\varepsilon \in \{ 4,8,16 \}$, and the number of iterations is calculated as $n = \min \{ \varepsilon+4, \lfloor 1.25\varepsilon \rfloor \} $. The attacks are then denoted by FGSM$_{\varepsilon}$ and I-FGSM$_{\varepsilon}$ to indicate the perturbation magnitude used, and a subscript $ll$ is added when an targeted attack with the least likely class as target is performed. 
For PGD, ALMA prox and DAG, we used the implementations and default parameter settings presented in \cite{Rony2022}. For these three attacks, we consider the untargeted as well as the targeted version choosing the semantic segmentation of the Cityscapes training image ``aachen\_000000\_000019'' as target.
We denote the targeted version as PGD$^{\mathit{tar}}$, ALMA$^{\mathit{tar}}$ and DAG$^{\mathit{tar}}$, respectively. 
For the DAG attack, we construct two more targets, i.e., predicting all car or pedestrian pixels as street, denoted by DAG$^{\mathit{tar}}_{\mathit{car}}$ and DAG$^{\mathit{tar}}_{\mathit{ped}}$.
For the two attacks, SSMM and DNNM, that are based on the universal perturbation, we follow the setting of \cite{Metzen2017} and choose a fixed segmentation of the Cityscapes dataset as target (the same as for the other targeted attacks) and delete the class person for DNNM.

\begin{figure}[t]
    \centering
    \subfloat[][Prediction]{\includegraphics[trim=0 0 0 0,clip,width=0.244\textwidth]{
    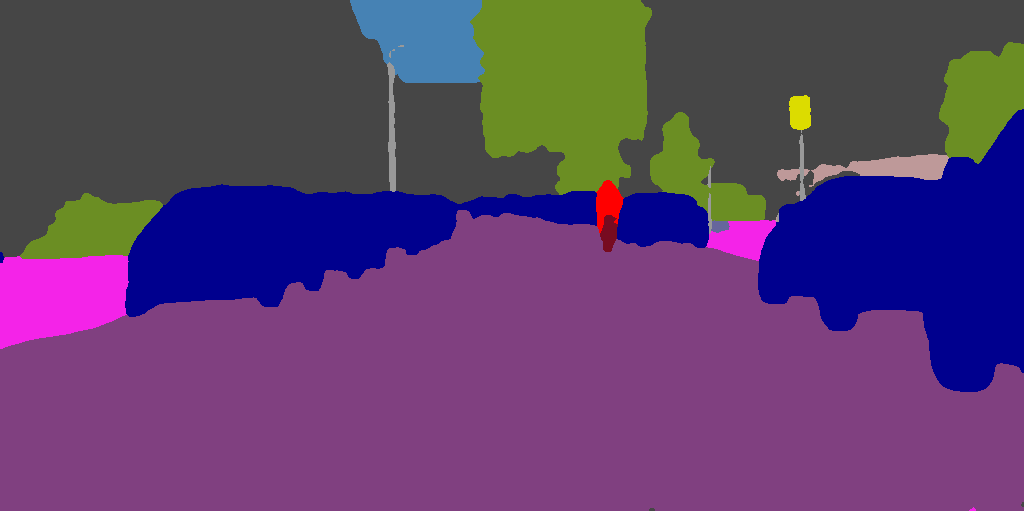} }
    \subfloat[][Heatmap of (a)]{\includegraphics[trim=0 0 0 0,clip,width=0.244\textwidth]{
    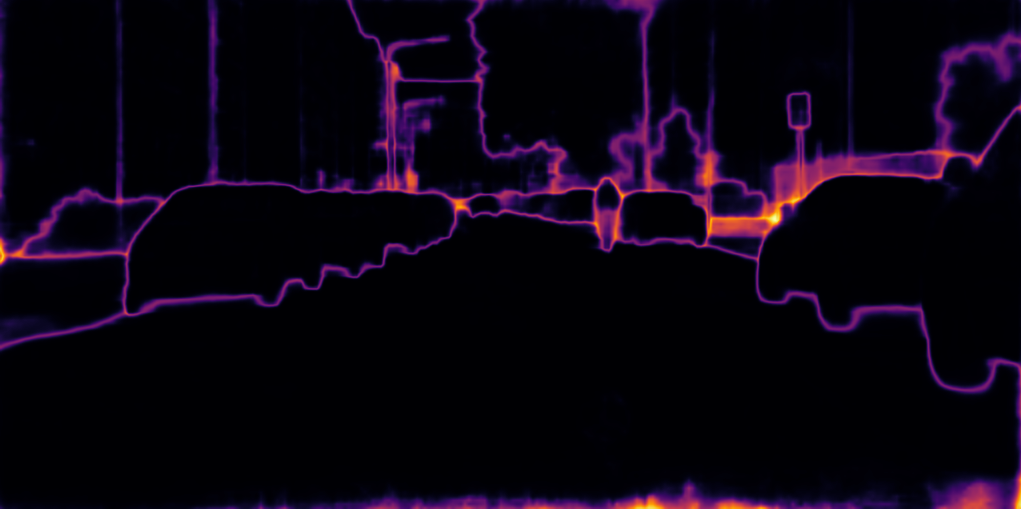} } 
    \subfloat[][FGSM$_{16}$]{\includegraphics[trim=0 0 0 0,clip,width=0.244\textwidth]{
    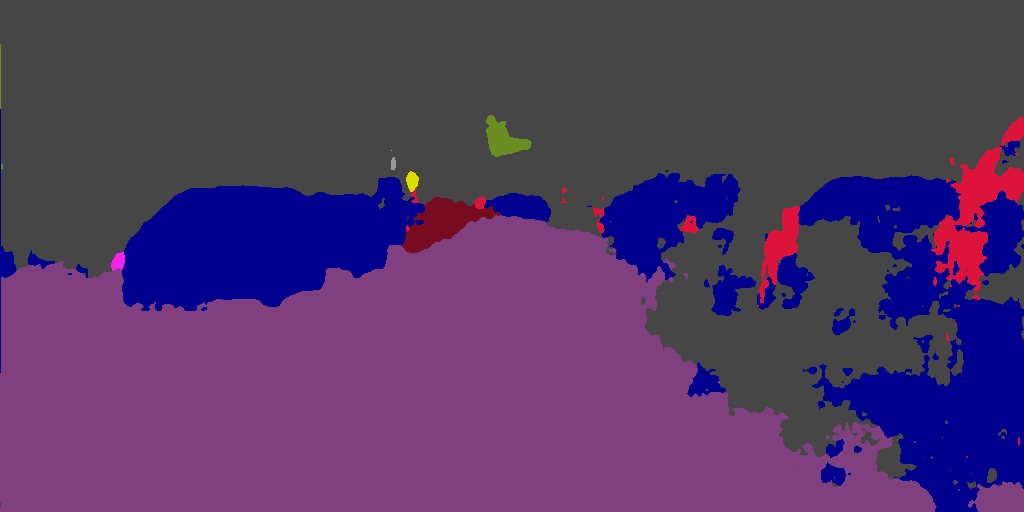} }
    \subfloat[][Heatmap of (c)]{\includegraphics[trim=0 0 0 0,clip,width=0.244\textwidth]{
    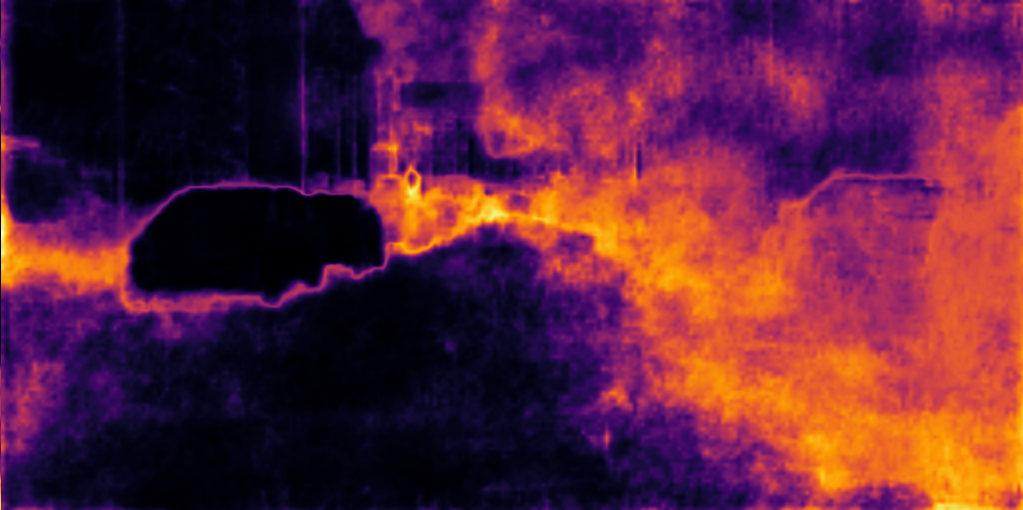} } \\
    \vspace{-2ex}
    \subfloat[][DAG]{\includegraphics[trim=0 0 0 0,clip,width=0.244\textwidth]{
    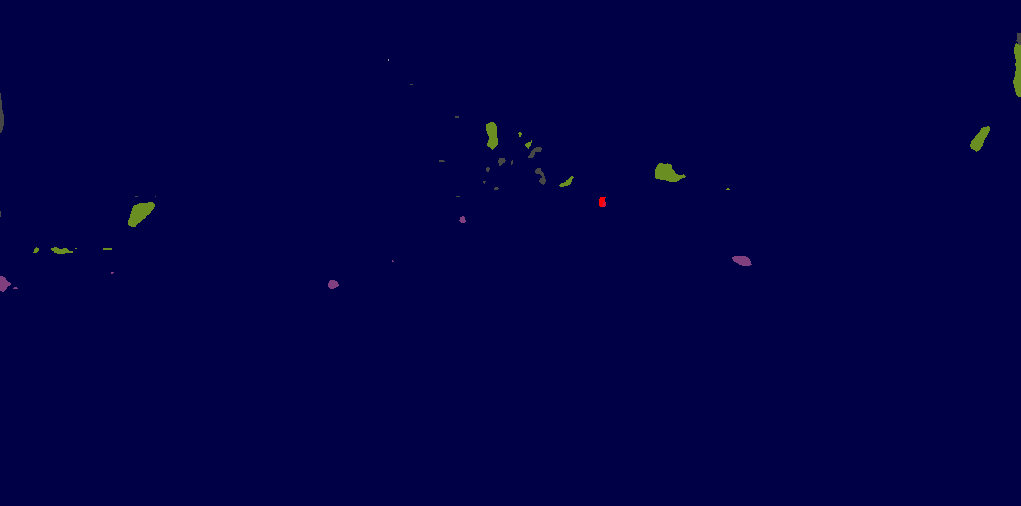} }
    \subfloat[][Heatmap of (e)]{\includegraphics[trim=0 0 0 0,clip,width=0.244\textwidth]{
    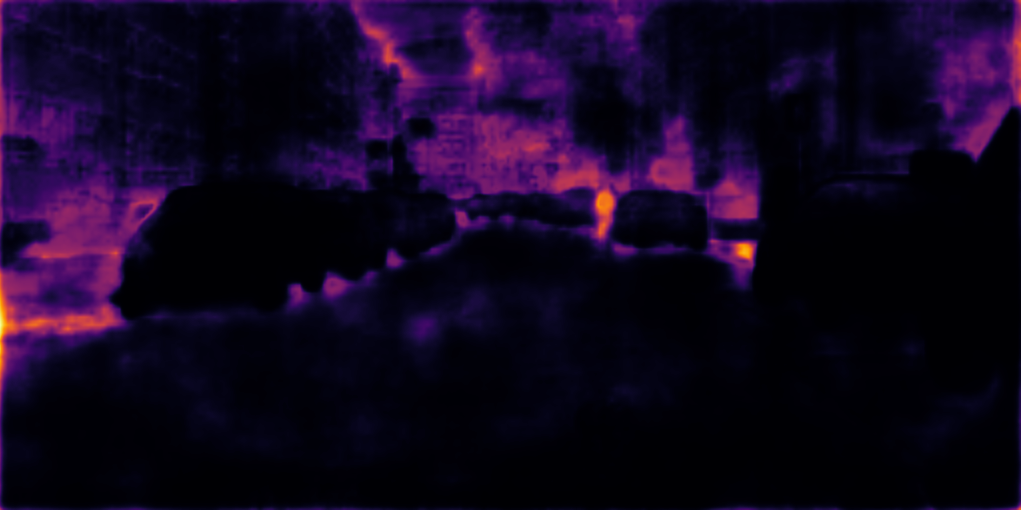} } 
    \subfloat[][I-FGSM$^{ll}_8$]{\includegraphics[trim=0 0 0 0,clip,width=0.244\textwidth]{
    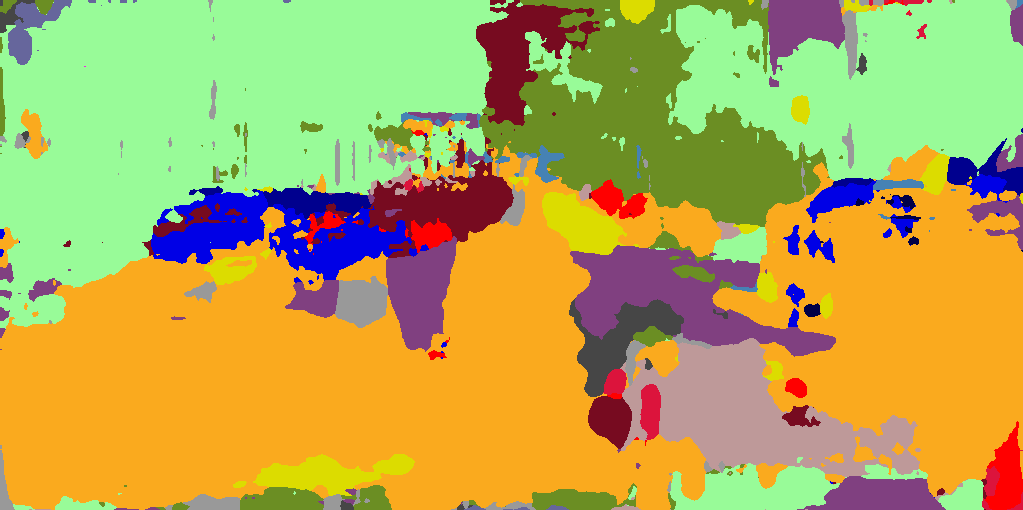} }
    \subfloat[][Heatmap of (g)]{\includegraphics[trim=0 0 0 0,clip,width=0.244\textwidth]{
    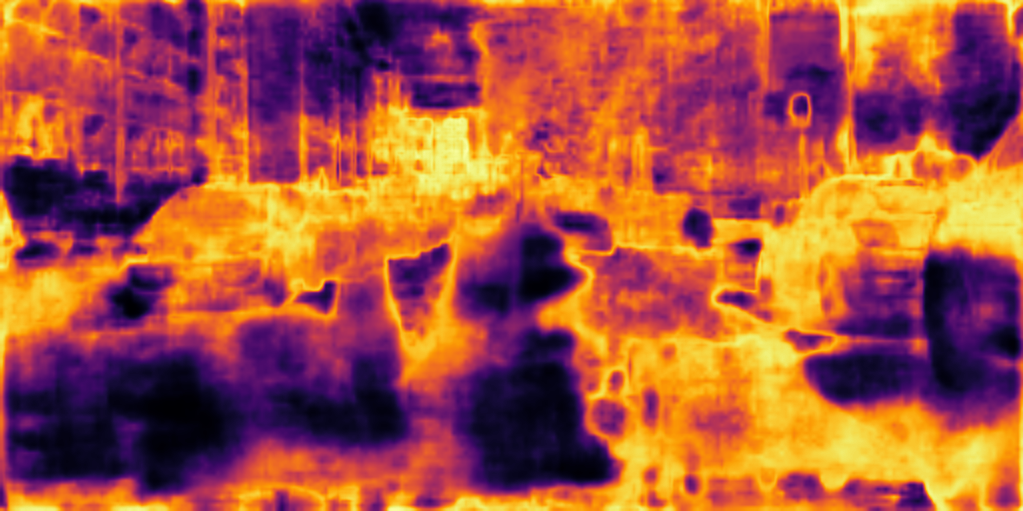} } \\
    \vspace{-2ex}
    \subfloat[][ALMA$^{\mathit{tar}}$]{\includegraphics[trim=0 0 0 0,clip,width=0.244\textwidth]{
    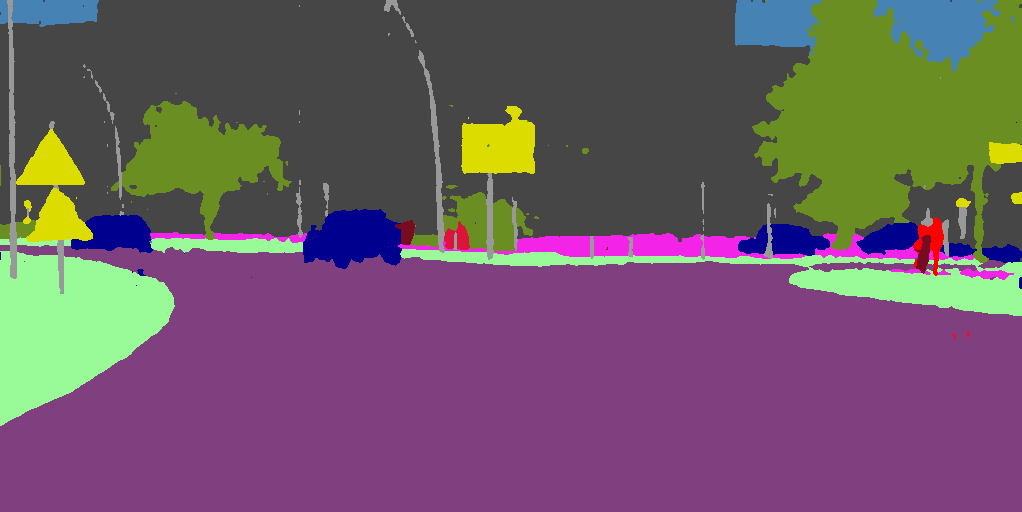} }
    \subfloat[][Heatmap of (i)]{\includegraphics[trim=0 0 0 0,clip,width=0.244\textwidth]{
    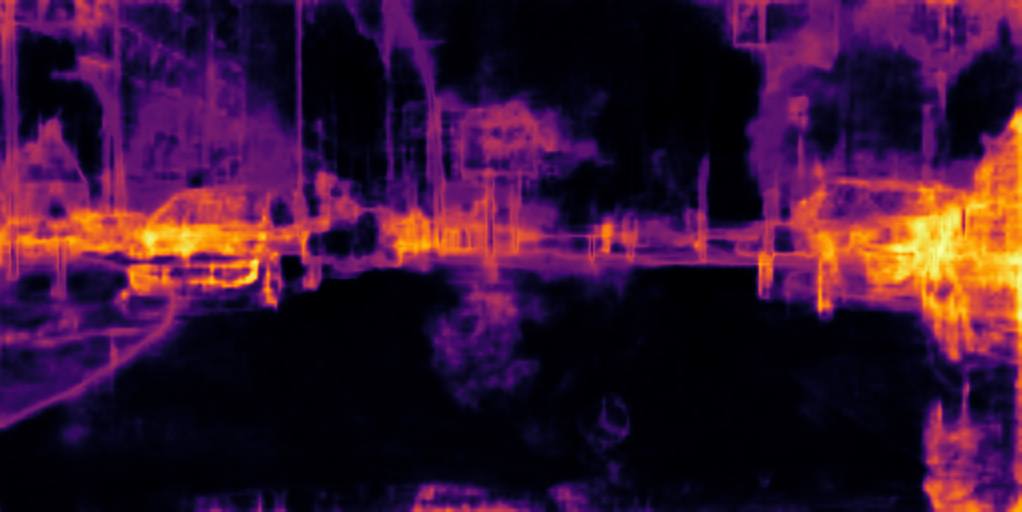} }
    \subfloat[][DNNM]{\includegraphics[trim=0 0 0 0,clip,width=0.244\textwidth]{
    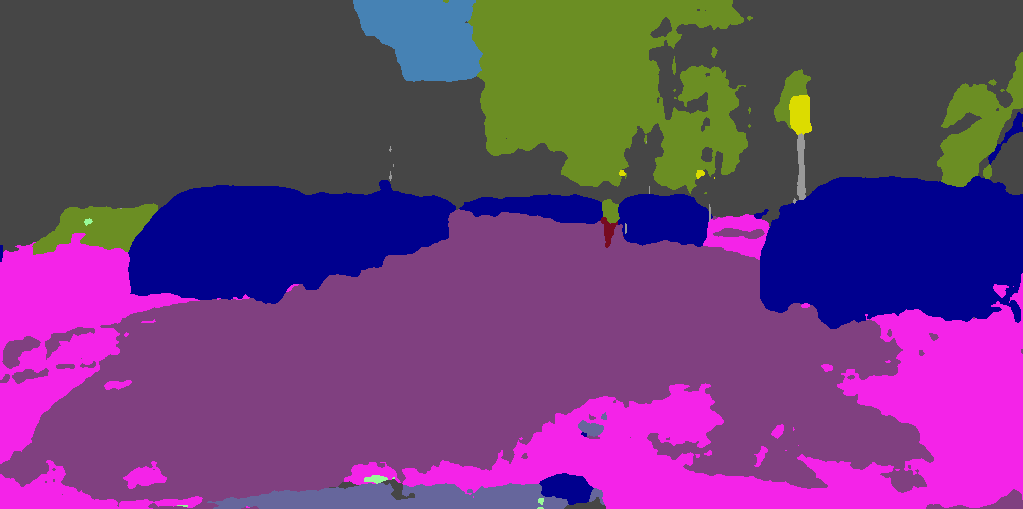} }
    \subfloat[][Heatmap of (k)]{\includegraphics[trim=0 0 0 0,clip,width=0.244\textwidth]{
    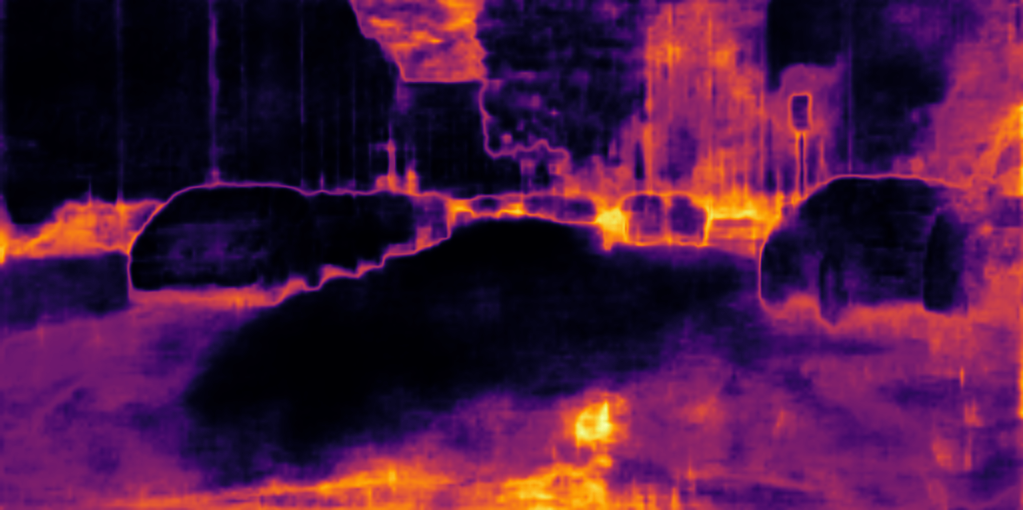} }
    \caption{Semantic segmentation prediction for clean (a) and perturbed image generated by different attacks (c)-(l) with corresponding entropy heatmaps.} 
    \label{fig:preds_adv}
\end{figure}

In our experiments, we use a model zoo\footnote{\url{https://github.com/open-mmlab/mmsegmentation}} with it's pre-trained models to run the attacks. In general, the Cityscapes images are resized to $512 \times 1024$ as this dataset provides high-resolution images ($1024 \times 2048$ pixels) which need a large amount of memory to run a full backward pass for the generation of adversarial samples. 
A few examples of different adversarial attacks applied to the DeepLabv3+ network are shown in Fig.~\ref{fig:preds_adv}.
%
%
\paragraph{Evaluation Metrics.} 
To access the strength of the adversarial attack generating methods, we consider the attack pixel success rate (APSR, \cite{Rony2022}). This metric measures falsely (and not correct) predicted pixels and thus indicates the opposite of the accuracy.

For accessing the detection performance of the proposed uncertainty-based approaches, we take different metrics into account. 
Recall, that the detection models provide a probability $d(x)$ per image of being benign (and not attacked) and classify an image as an attack if $d(x)$ is smaller than a  predefined threshold $\tau$. In our experiments, we choose $40$ different values equally spaced in $[0,1]$ for the threshold $\tau$.
Based on this, we calculate three different evaluation metrics. Firstly, we use the optimal averaged detection accuracy (ADA) computed as $\ada = \max_{\tau \in [0,1]} \text{ADA}(\tau)$, where each ADA value depends on a threshold $\tau$ and defines the proportion of images that are classified correctly as benign or adversary.
Secondly, we use the area under the receiver operating characteristic curve ($\auc$) to obtain a metric which does not depend on the threshold. 
Thirdly, we consider the true positive rate while fixing the false positive rate on benign images to $5\%$ ($\tpr$) as safety-critical measure.
%
%
\paragraph{Classification Models.} 
In our experiments, we refer to the single-feature, mean entropy-based classification as \emph{Entropy}, the standard one-class support vector machine as \emph{OCSVM}, and the outlier detection method from \cite{Rousseeuw1999} as \emph{Ellipse}. 
For training the logistic regression model, we use benign data and adversarial examples from a targeted FGSM attack with a noise magnitude of $2$ as perturbed data, under the assumption that using data from an attack with minimal perturbation strength could be advantageous for detecting more subtle attacks. 
 We denote this regression classifier by \emph{CrossA}.
We proceed analogously for the shallow neural network classifier which is fed with the complete heatmaps as input. We refer to this method as \emph{Heatmap}. We evaluate all detection models using 5-fold cross-validation. 

Note, in general, we do not have knowledge of the adversarial example generation process used by the attacker. Therefore, we use adversarial examples stemming from only one adversarial attack to train the classifier and then test its robustness against all other attack types. 
Of cause, the performance of the classifier could be improved by including a more diverse set of adversarial examples into the training set, but we decided to stick to this kind of "worse-case performance" for our evaluation (resembling the setting where the attacker uses an attack not know by the model).
%
%
\subsection{Numerical Results}\label{sec:exp_results}
\begin{figure}[t]
    \centering
    \includegraphics[width=0.99\textwidth]{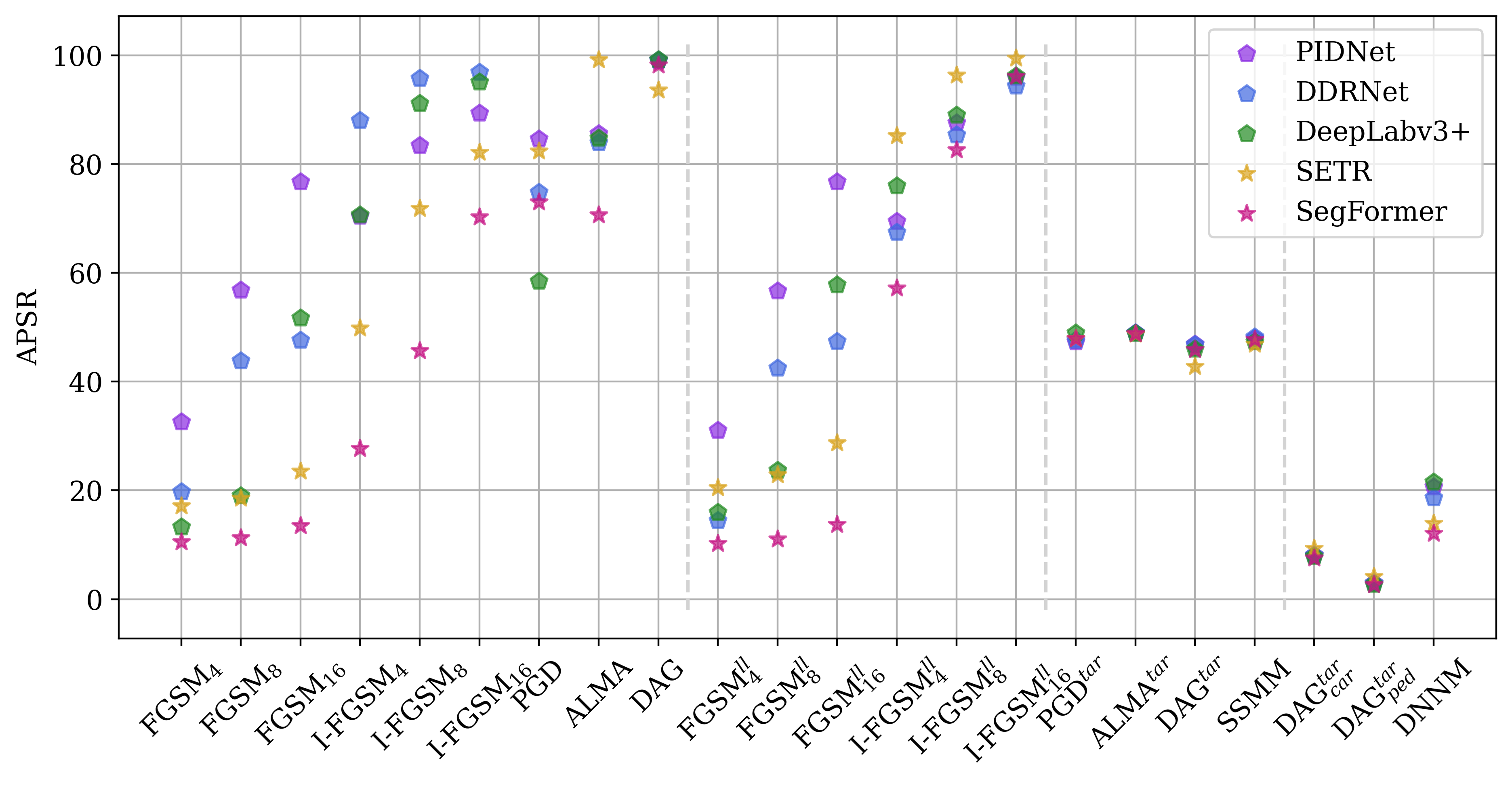} 
    \caption{APSR results for the Cityscapes dataset perturbed by various attacks. } 
    \label{fig:apsr}
\end{figure}

In the following, we start with investigating the success performance of the adversarial attacks and then evaluate the attack detection performance of the proposed approach.
%
%
\paragraph{Performance of the Adversarial Attacks.} 
The APSR results for various attacks on the Cityscapes dataset are given in Fig.~\ref{fig:apsr}.
The figure can be separated in 4 parts (indicated by dashed lines). 
The first part contains all untargeted attacks, the second part attacks with the least likely class as the target, the third part attacks with a static image as the target, and the last part attacks in which a class is deleted. 
In general, for all variations of the FGSM attack (non-iterative vs. iterative, untargeted vs. targeted) the APSR value increases as the perturbation strength magnitude increases. Furthermore, the I-FGSM outperforms the FGSM due to its iterative procedure involving multiple perturbation steps. 
For the untargeted case, PGD achieves similar results to I-FGSM for most models, as the nature of the two attacks is similar. Moreover, both attacks specifically developed for the semantic segmentation task, ALMA and DAG, result in strong APSR scores. 
For the targeted attacks with static image, the target is the segmentation of a (randomly) selected image from the Cityscapes dataset.
As illustrated in Fig.~\ref{fig:preds_adv} (a) and (i), the correct and target classes of both, the clean and perturbed images, overlap in several regions, such as the street and the buildings, which is characteristic of street scenes. This overlap may account for the relatively low ASPR values, which hover around $50\%$.
For the targeted attacks shown in the last part of Fig.~\ref{fig:apsr}, the APSR scores are relatively low as only a single class is targeted for deletion, leaving most parts of the targeted segmentation map unchanged.

\begin{figure}[t]
    \centering
    \includegraphics[width=0.99\textwidth]{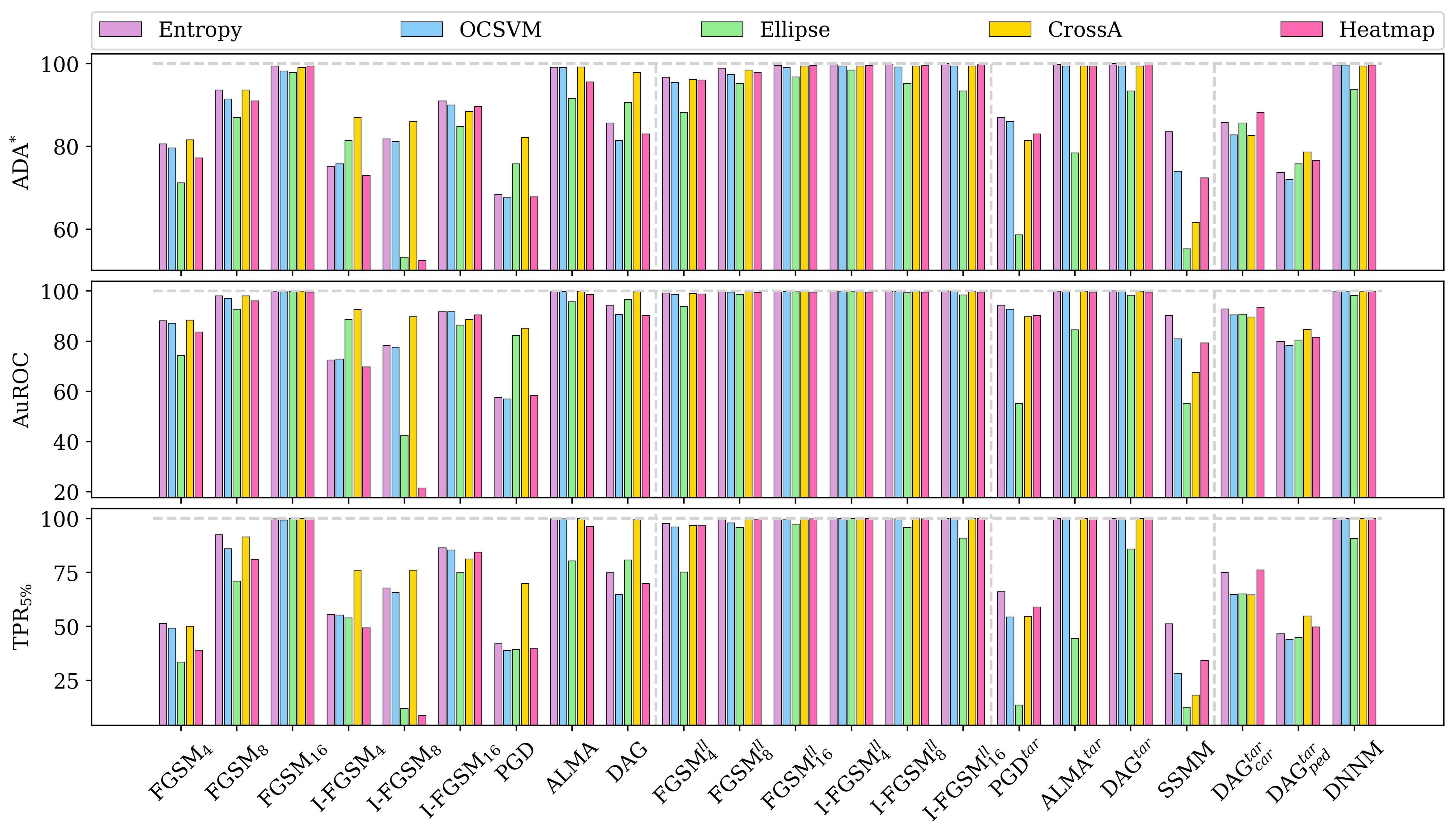} 
    \caption{Detection performance results for the DeepLabv3+ network.} 
    \label{fig:nr_deep}
\end{figure}

Comparing the APSR values between the models, it is striking that the transformer-based networks are often more robust to various attacks than the convolutional networks. For the convolutional models, we observe that there is no clear winner in terms of robustness, i.e., no network always obtains smaller APSR values compared to the other networks across all attacks. However, there is a trend among the two transformers that the SegFormer is more robust against adversarial attacks than SETR. 
To summarize, most attacks achieve high APSR values for the different network architectures and strongly change the prediction. Thus, the detection of such attacks is highly relevant.
%
%
\paragraph{Performance of the Uncertainty-based Detection Approach.} 
The detection results for the DeepLabv3+ are given in Fig.~\ref{fig:nr_deep}.
We observe that the detection performance is comparatively higher for greater perturbation magnitudes in FGSM attacks. This can be attributed to weaker attacks causing changes in predictions for a fewer number of pixels, making them harder to detect.
The detectors do not perform well on adversarial examples generated by untargeted I-FGSM, despite the attack's strength. 
Inspection of these examples reveals that during segmentation, only a few classes are predicted, often with low uncertainty for large connected components. This makes it difficult to distinguish between benign and perturbed data.
The targeted FGSM attacks are detected more effectively than untargeted attacks. This may be because targeting the most unlikely class leads to more significant changes in the uncertainty measures used as features, see Fig.~\ref{fig:preds_adv} (g).
In the targeted attacks with static image, it is noticeable that the detection performance for PGD$^\mathit{tar}$ and SSMM is comparatively smaller than for ALMA$^\mathit{tar}$ and DAG$^\mathit{tar}$. The latter two attacks lead the DeepLabv3+ to predict the whole target image rather well but with high uncertainty in some areas (see Fig.~\ref{fig:preds_adv} (j)), while PGD$^\mathit{tar}$ and SSMM predict large areas of the image accurately and reliably and have only smaller uncertain areas.  
The detection capability for targeted attacks with the aim of deleting a class stands out positively. Even though this attacks are especially challenging to detect, as only a few pixels are changed while most remain unchanged, AuROC values between $80$ and $100$ are reached.
In general, the basic Entropy method already achieves high scores, however, the supervised methods (CrossA and Heatmap) frequently achieve the highest results.

\begin{figure}[t]
    \centering
    \includegraphics[trim=0 325 0 0,clip,width=0.99\textwidth]{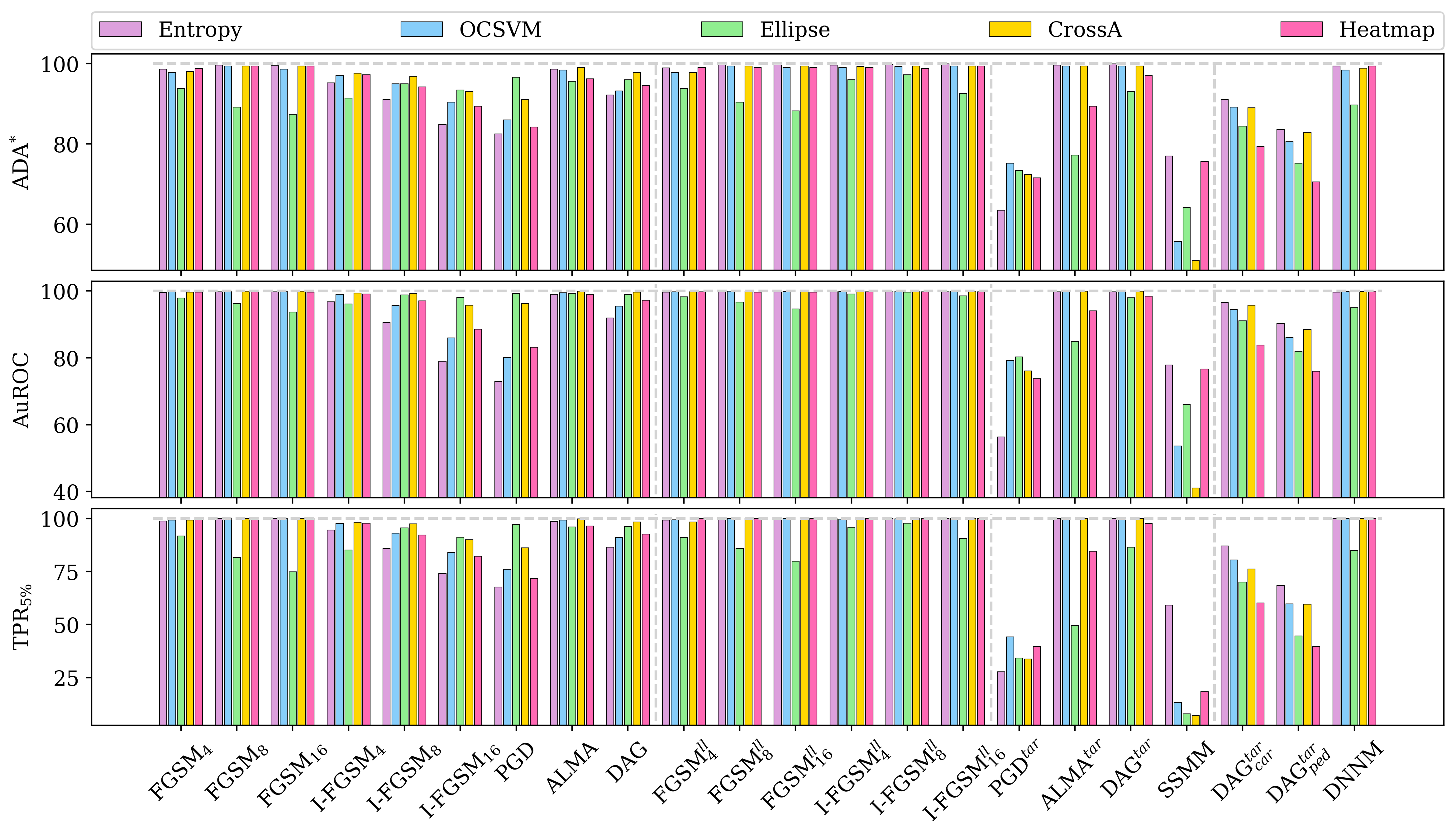} 
    \includegraphics[trim=0 325 0 30,clip,width=0.99\textwidth]{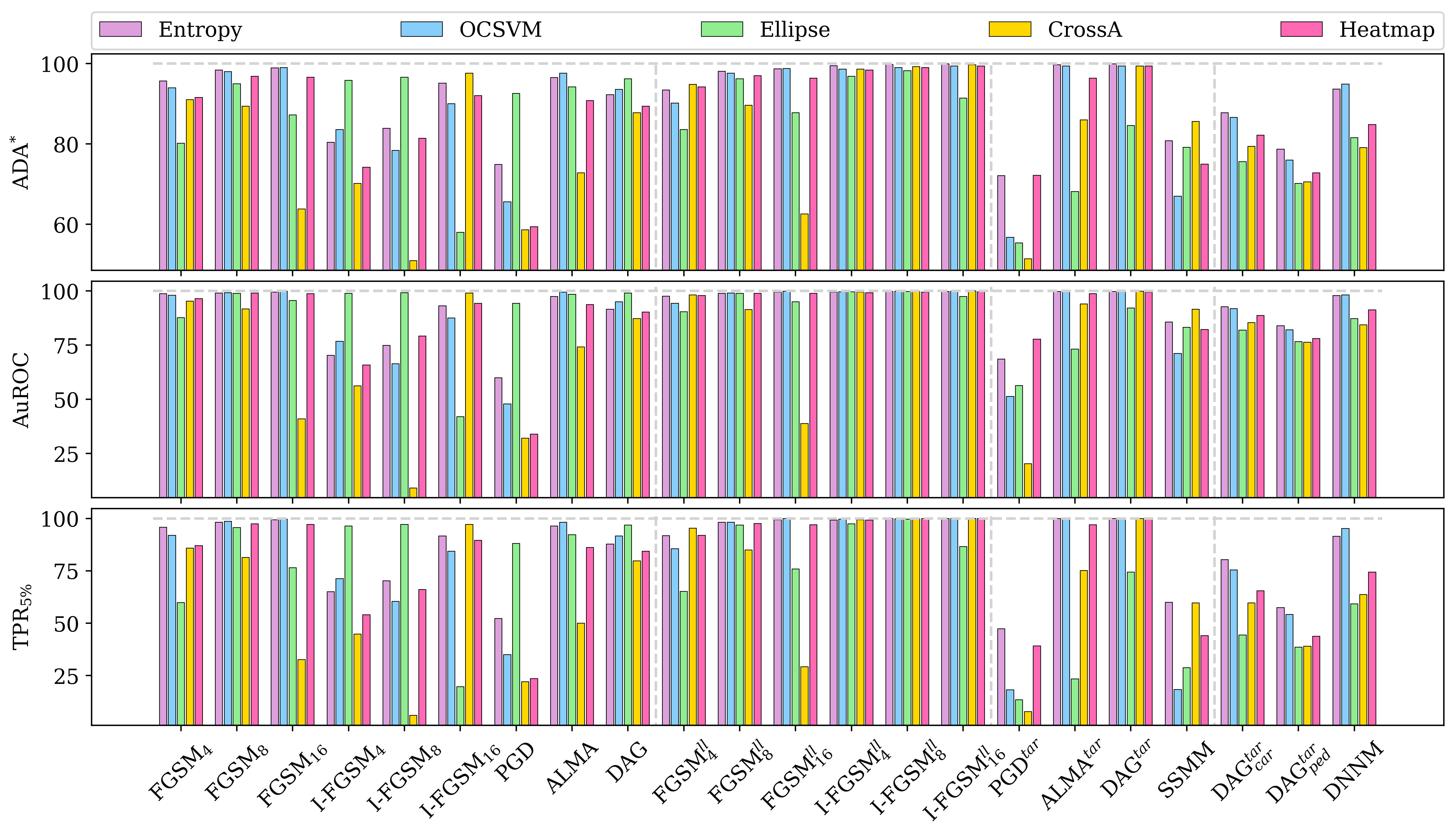}
    \includegraphics[trim=0 0 0 426,clip,width=0.99\textwidth]{figs/fig_res_ci_ddrn.png}
    \caption{Detection performance results for the PIDNet (\emph{top}) and the DDRNet (\emph{bottom}) network.} 
    \label{fig:nr_pidn_ddrn}
\end{figure}

\begin{figure}[t]
    \centering
    \includegraphics[trim=0 325 0 0,clip,width=0.99\textwidth]{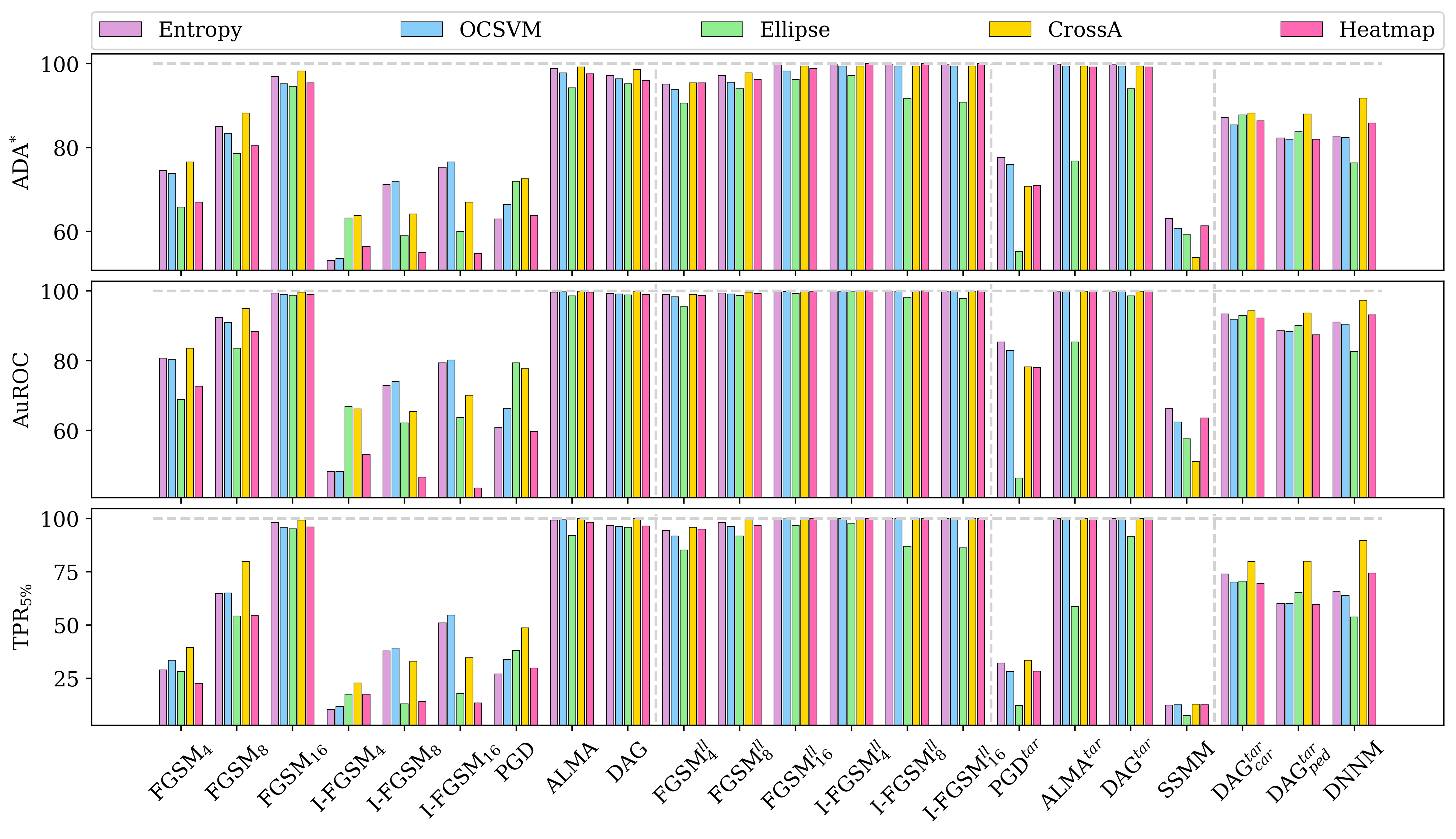} 
    \includegraphics[trim=0 325 0 30,clip,width=0.99\textwidth]{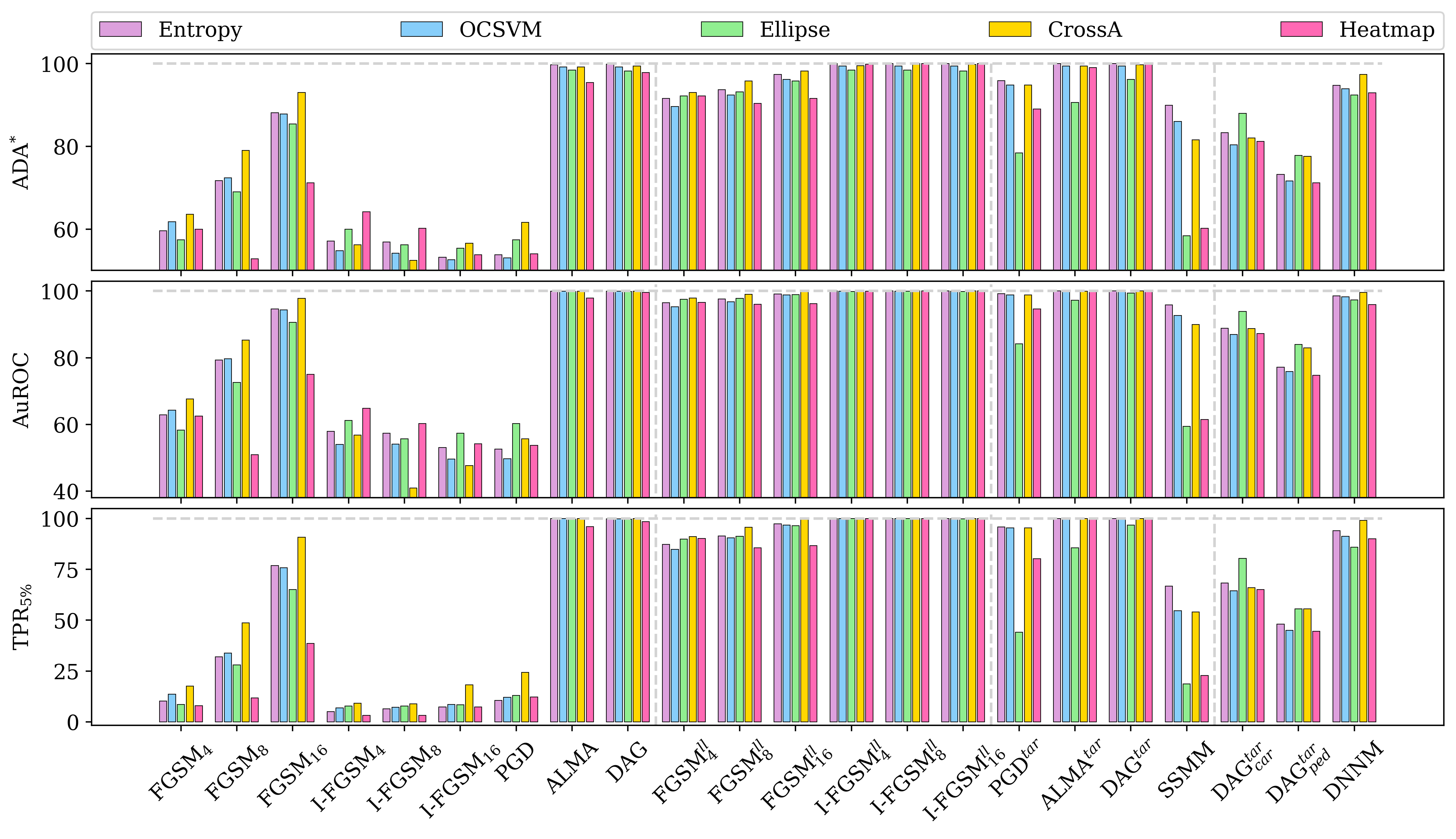}
    \includegraphics[trim=0 0 0 426,clip,width=0.99\textwidth]{figs/fig_res_ci_segf.png}
    \caption{Detection performance results for the SETR (\emph{top}) and the SegFormer (\emph{bottom}) network.} 
    \label{fig:nr_setr_segf}
\end{figure}

For the analysis of the detection performance for the other segmentation models, we will focus on the $\ada$ value for the sake of more compact representation, since the three evaluation metrics show the same trends for the different attacks and detection methods.
The detection results for the PIDNet and DDRNet are shown in Fig.~\ref{fig:nr_pidn_ddrn}.
We observe a similar behavior of the detection methods across the different attacks.
Overall, the detection capability for attacks on the PIDNet is higher than for the DeepLabv3+, as the PIDNet is more vulnerable to attacks leading to larger changes in segmentation predictions (as shown by the larger APSR values in Fig.~\ref{fig:apsr}).
In the DDRNet, it is striking that the unsupervised classifiers outperform the supervised methods, which indicates that the FGSM$^{ll}_2$ does not provide generalizable information on the other attacks.
Finally, the detection results for transformer SETR and SegFormer are given in Fig.~\ref{fig:nr_setr_segf}.
As shown by the results on attack strength depicted in Fig.~\ref{fig:apsr}, the APSR values for the transformer-based networks are often lower, i.e., the networks are more robust against attacks. This characteristic is also reflected in the detection performance, as the values are comparatively smaller. 
In this case, it is useful to have already seen perturbed data when detecting, since the supervised methods (in particular CrossA) clearly outperform the other methods.
In general, our experiments show the huge potential of studying uncertainty information for the effective detection of adversarial attacks in semantic segmentation.
%
%
%
\section{Conclusion}\label{sec:conc}
We have proposed an uncertainty-based approach for the detection of adversarial attacks on semantic segmentation models. The motivation is that entropy-based uncertainty information exhibits different behavior in benign versus adversary perturbed images. We conducted our tests on a wide range of adversarial attacks as well as state-of-the-art segmentation networks (like transformers). To perform the detection, we trained various classifiers (supervised as well as unsupervised) based on pixel-wise uncertainty measurements aggregated over the images or on the entire heatmaps (i.e., all pixel-wise uncertainty estimates). 
Our experiments show that transformers as well as convolutional networks are vulnerable to adversarial attacks. However, the proposed adversarial example detection approach serves as a strong defense mechanism. It achieves an average optimal detection accuracy rate of $89.36\%$ over the different attacks and segmentation models. 
Moreover, it is lightweight, i.e., it can be seen as a post-processing approach that requires no modifications to the original model or insight into the attacker's method for creating adversarial examples. 
In conclusion, the strong detection performance and the lightweight nature of our methodmake it a strong baseline  for developing more elaborated uncertainty-based detection methods in future.

\begin{credits}
\subsubsection{\ackname} This work is supported by the Deutsche Forschungsgemeinschaft (DFG, German Research Foundation) under Germany’s Excellence Strategy – EXC-2092 CASA – 390781972.

\end{credits}
%
%
%
\bibliographystyle{splncs04}
\bibliography{mybibliography}


\end{document}

%% file: figs/method.tex
\begin{tikzpicture}

\draw[fill=lavenderblush,rounded corners] (0, 0) rectangle (2.5,1.2) {};
\node at (1.25,0.8) {input image};
\node at (1.25,0.4) {$x$};

\draw [-Latex,thick] (2.6,0.6) -- (3.6,0.6);

\draw[fill=honeydew,rounded corners] (3.7, 0) rectangle (6.2,1.2) {};
\node at (4.95,0.8) {segmentation};
\node at (4.95,0.4) {network $f$};

\draw [-Latex,thick] (6.3,0.6) -- (7.3,0.6);

\draw[fill=honeydew,rounded corners] (7.4, 0) rectangle (9.9,1.2) {};
\node at (8.65,0.8) {softmax};
\node at (8.65,0.4) {probs $f(x;w)$};

\draw [-Latex,thick] (10.0,0.6) -- (11.0,0.6);

\draw[fill=honeydew,rounded corners] (11.1, 0) rectangle (13.6,1.2) {};
\node at (12.35,0.8) {semantic seg-};
\node at (12.35,0.4) {mentation $\hat{y}^x$};

\draw [-Latex,thick,dotted] (4.95,-0.9) -- (4.95,-0.1);

\draw[fill=bananamania,rounded corners] (3.7, -2.2) rectangle (6.2,-1.0) {};
\node at (4.95,-1.4) {adversarial};
\node at (4.95,-1.8) {attacker};








\draw[fill=aliceblue,rounded corners] (7.4, -3.0) rectangle (9.9, -1.8) {};
\node at (8.65,-2.2) {detector};
\node at (8.65,-2.6) {$d(x)$};

\draw [-Latex,thick] (10.0,-2.3) -- (11.5,-1.6);
\draw [-Latex,thick] (10.0,-2.5) -- (11.5,-3.2);

\node[rotate=22] at (10.7,-1.5) {$d(x) \geq \tau$};
\node[rotate=-22] at (10.7,-3.3) {$d(x) < \tau$};

\draw[fill=aliceblue,rounded corners] (11.6, -2) rectangle (14.1,-1.2) {};
\node at (12.85,-1.6) {benign};

\draw[fill=aliceblue,rounded corners] (11.6, -2.8) rectangle (14.1,-3.6) {};
\node at (12.85,-3.2) {perturbed};

\draw [-Latex,thick, dashed] (8.65,-0.1) -- (8.65,-1.7);
\node at (10.1,-0.7) {heatmaps/features};

\end{tikzpicture}